\documentclass[runningheads]{llncs}
\usepackage[T1]{fontenc}
\usepackage{graphicx}
\usepackage{marvosym}
\begin{document}
\title{Unsupervised Multimodal 3D Medical Image Registration with Multilevel Correlation Balanced Optimization}
\author{Jiazheng Wang\inst{1, 2} \and
Xiang Chen\inst{1, 2} \and
Yuxi Zhang\inst{1, 2} \and
Min Liu\inst{1, 2}\textsuperscript{(\Letter)} \and
Yaonan Wang\inst{1, 2} \and
Hang Zhang\inst{3}}
\authorrunning{J. Wang et al.}
\institute{College of Electrical and Information Engineering, Hunan University, Changsha, Hunan, China \and
National Engineering Research Center of Robot Visual Perception and Control Technology, Hunan University, Changsha, Hunan, China \\
\email{\{wjiazheng, xiangc, hnuzyx, liu\_min, yaonan\}@hnu.edu.cn}
\and Cornell University, USA \\
\email{\{hz459\}@cornell.edu}}
\maketitle              % typeset the header of the contribution
\begin{abstract}

Surgical navigation based on multimodal image registration has played a significant role in providing intraoperative guidance to surgeons by showing the relative position of the target area to critical anatomical structures during surgery. However, due to the differences between multimodal images and intraoperative image deformation caused by tissue displacement and removal during surgery, effective registration of preoperative and intraoperative multimodal images faces significant challenges. To address the multimodal image registration challenges in Learn2Reg 2024, an unsupervised multimodal medical image registration method based on multilevel correlation balanced optimization (MCBO) is designed to solve these problems. First, the features of each modality are extracted based on the modality independent neighborhood descriptor, and the multimodal images are mapped to the feature space. Second, a multilevel pyramidal fusion optimization mechanism is designed to achieve global optimization and local detail complementation of the deformation field through dense correlation analysis and weight-balanced coupled convex optimization for input features at different scales. For preoperative medical images in different modalities, the alignment and stacking of valid information between different modalities is achieved by the maximum fusion between deformation fields. Our method focuses on the ReMIND2Reg task in Learn2Reg 2024, and to verify the generality of the method, we also tested it on the COMULIS3DCLEM task. Based on the results, our method achieved second place in the validation of both two tasks. The code is available at https://github.com/wjiazheng/MCBO.

\end{abstract}
\section{Introduction}

Medical image registration has been an important topic in the field of medical image analysis, and many significant methods \cite{ref_article4,ref_article5,ref_article6,ref_article8} have driven the development of medical image registration tasks. Deep learning-based medical image registration methods \cite{ref_article7} generally involve long and complex learning processes, and often struggle to achieve accurate estimation for multimodal, large-deformation data and general usability for extensive tasks. The Learn2Reg 2024 subchallenge, ReMIND2Reg, is a multimodal medical image registration task oriented to preoperative ultrasound and intraoperative magnetic resonance imaging (MRI), which is characterized as unlabeled, large deformation, and low feature distinctness. Aiming at the above characteristics, inspired by \cite{ref_proc1,ref_article1}, an unsupervised multimodal medical image registration method based on multilevel correlation balanced optimization (MCBO) has been proposed, which can quickly achieve effective multimodal medical image registration using only a small number of learning and optimization procedures.

\begin{figure}
\includegraphics[width=\textwidth]{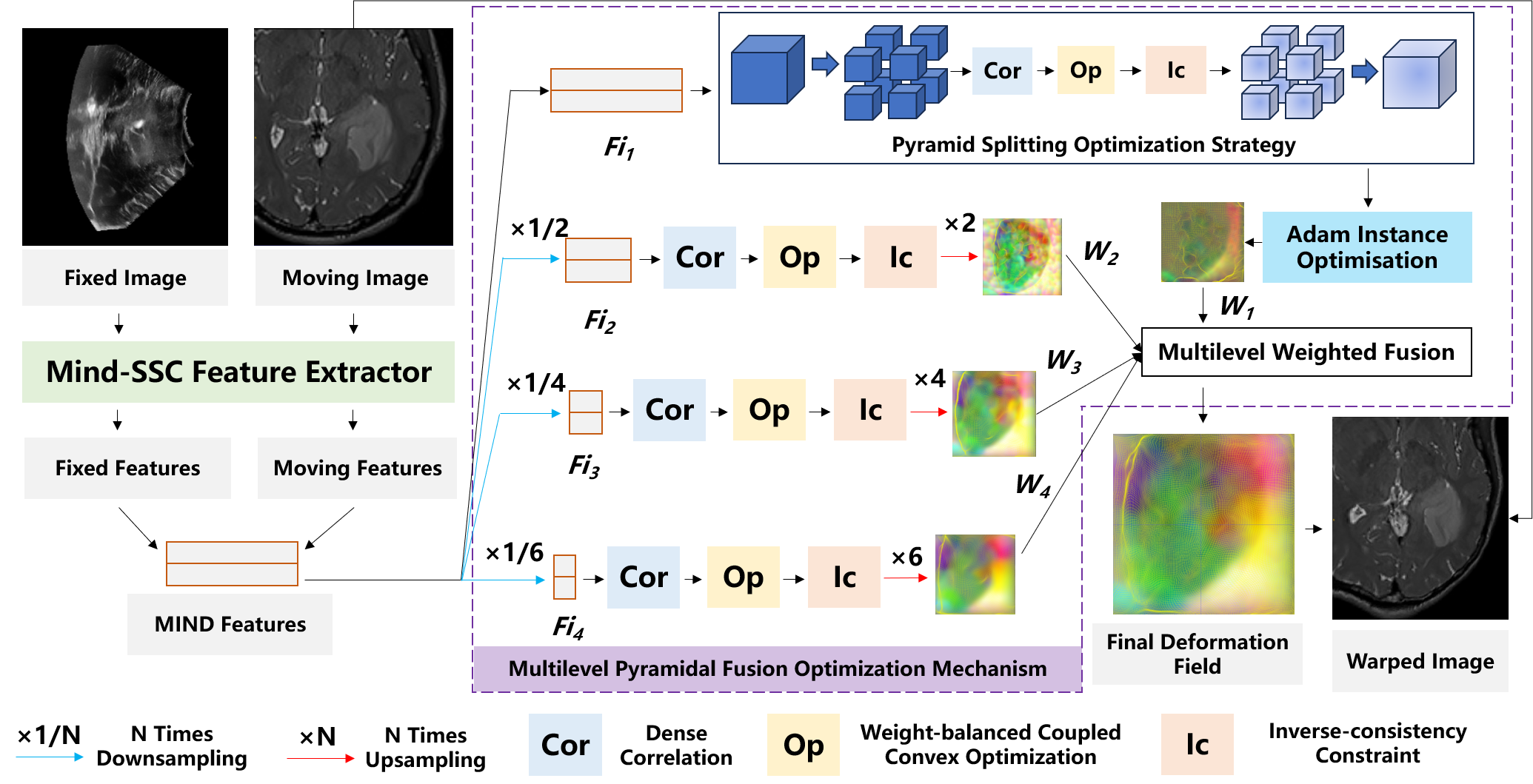}
\caption{The overall flow of the proposed MCBO method.} \label{fig1}
\end{figure}

\section{Methodology}

%a https://github.com/multimodallearning/convexAdam/tree/main
The proposed MCBO method is based on convexAdam \cite{ref_url1} with a series of improvements, which include (1) Introducing a weight-balancing term on coupled convex optimization to achieve smoother deformation optimization. (2) A multilevel pyramidal fusion optimization mechanism is designed to achieve refinement of the dense deformation field by fusing the optimization results of different scales. (3) For multimodal preoperative medical images in the task, the alignment and stacking of valid information between different modalities is achieved through the maximum fusion of deformation fields.

The overall flow of the proposed MCBO method is shown in Fig.~\ref{fig1}. The moving image and the fixed image are inputted and then the modal-independent features of the images are first obtained by Mind-SSC Feature Extractor \cite{ref_article1}. The Mind-SSC feature extractor exploits the self-similarity of the partial area in the image to extract the unique structural information of the local neighborhood, which results in a highly consistent structural representation across modalities.

\begin{figure}
\includegraphics[width=\textwidth]{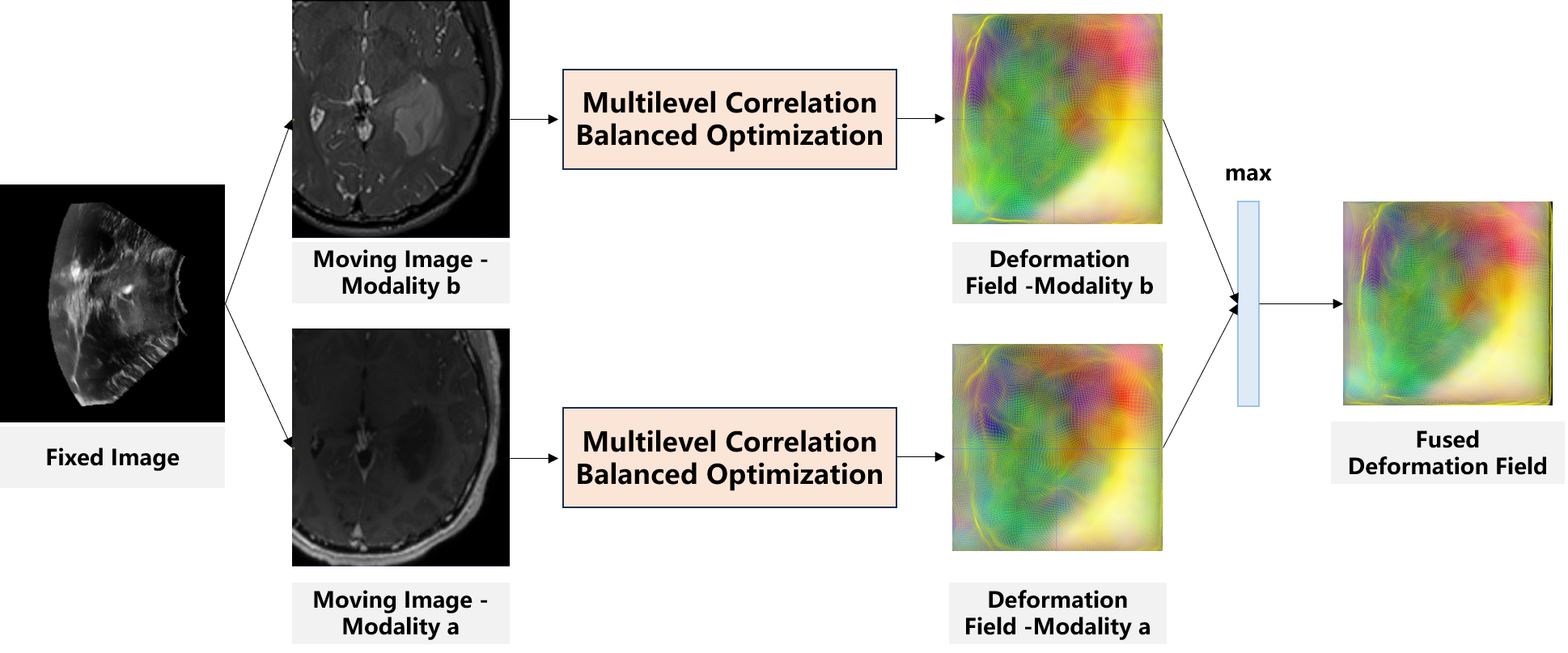}
\caption{The fusion of different modalities.} \label{fig2}
\end{figure}

The acquired features are subsequently downsampled by average pooling operations at different times $n$ ($n=1,2,4,6$) to obtain leveled features $Fi_{1}$ to $Fi_{4}$ as input for the multilevel pyramidal fusion optimization mechanism. For the input features of levels 2 to 4, the feature matrix is directly passed through the dense correlation layer, the weight-balanced coupled convex optimization layer, and the inverse-consistency constraint layer to obtain the initial deformation fields of each level. For the first level where the input features are of the original scale, a pyramid splitting optimization strategy is used to reduce the amount of computation during the optimization procedure and, at the same time, to enhance the ability to align the low-contrast features. Specifically, the original feature is first divided into eight parts by equally splitting along the H, W, D dimensions of the input feature respectively, and the same three-step process as the other levels is performed for each part of the feature separately, followed by splicing the feature in the original dimensions to get the initial deformation field of the first level. To ensure smoothness of the deformation field after splicing, an additional adam instance optimization \cite{ref_proc1} operation is performed on the output of the first level. Finally, the deformation fields at each level are summed up by multilevel weighted fusion using different weights, where the sum of the weights is 1, to obtain the final deformation field.

In particular, the acquired input features are first fed into the dense correlation layer to compute the sum-of-squared-differences (SSD) cost volume and the initial optimal displacements for each voxel. The large search space allows us to make an initial capture of the displacement for each voxel, even though some voxel points may have large deformations. The output of the dense correlation layer is alternately optimized for similarity and smoothness by iterations in the weight-balanced coupled convex optimization layer, and then the iterative optimization results are averaged to achieve global regularization with weight balancing. After that, an inverse consistency constraint layer \cite{ref_proc2} is introduced to minimize the difference between the forward and backward transformations to avoid incredible deformations. 
 
For moving images with multiple modalities, the alignment and stacking of valid information between different modalities is realized by the maximum fusion of deformation fields, and the detailed steps are shown in Fig.~\ref{fig2}.

\begin{figure}
\includegraphics[width=\textwidth]{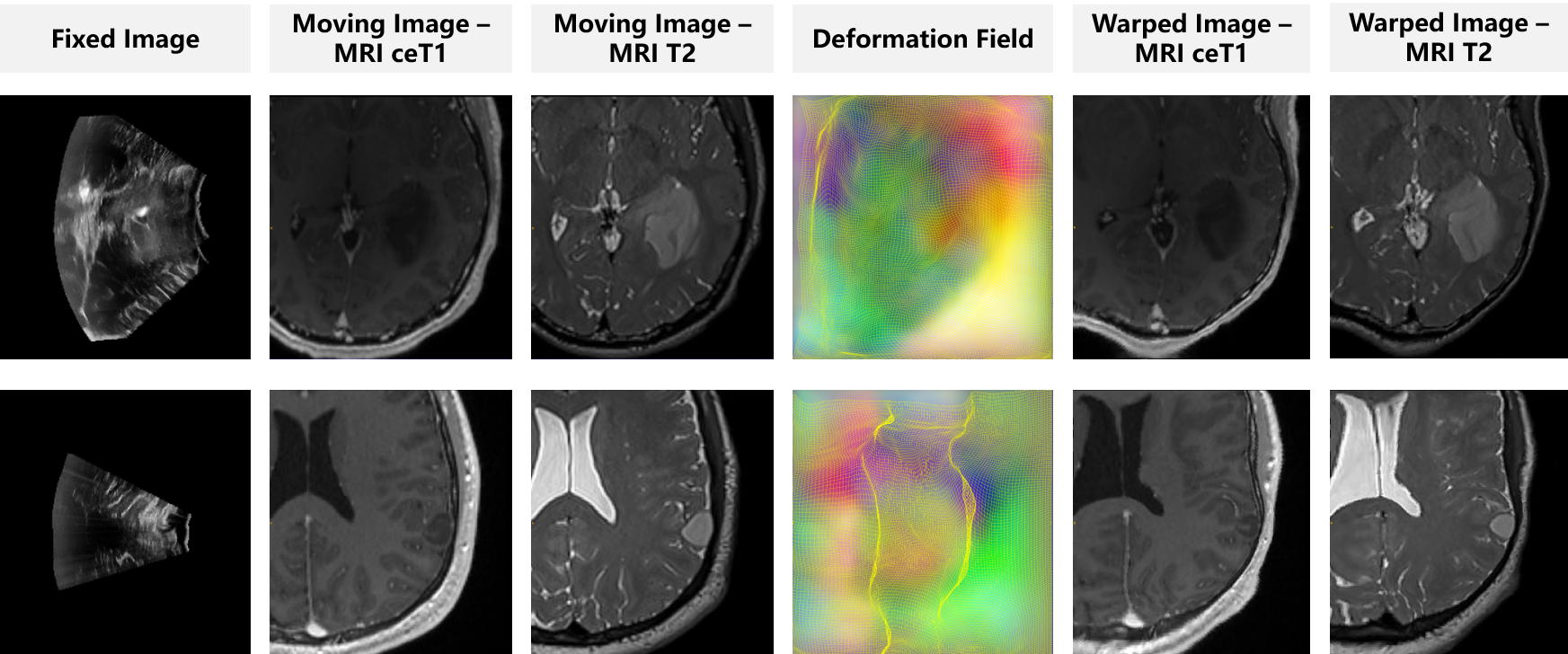}
\caption{Visualization results of ReMIND2Reg sub-challenge in Learn2Reg 2024.} \label{fig3}
\end{figure}

\section{Experiments and Results}

Our method focuses on the ReMIND2Reg sub-challenge task in Learn2Reg 2024, and to verify the generality of the method, we also tested it on the COMULIS3DCLEM sub-challenge task. The experimental setup of the method is slightly different in the two sub-challenges and the results demonstrate the effectiveness of our method.

\paragraph{ReMIND2Reg task.}
The goal of the ReMIND2Reg \cite{ref_article2} sub-challenge is to register preoperative MRI from multiple modalities (including ceT1 and T2) and intraoperative 3D ultrasound images. For this task, we set the weights of each level in Multilevel Weighted Fusion as 0.10, 0.27, 0.27, 0.36. Meanwhile, for the adam instance optimization operation performed in the first level, the number of iterations is set to 15, the smooth convolution kernel is set to 5, and the rest of the parameters are referred to the original settings\cite{ref_url1}. The experimental results are shown in Table~\ref{tab1}. The visualization of the multimodal image registration for this task is shown in Fig.~\ref{fig3}. In the validation phase, our method ranks in the second place.

\begin{table}
\centering
\caption{Results of ReMIND2Reg sub-challenge in Learn2Reg 2024.}\label{tab1}
\begin{tabular}{|c|c|}
\hline
  &  TRE(mm) \\
\hline
Initial & 3.727 $\pm$ 0.714 \\
ConvexAdam-Rigid &  2.773 $\pm$ 1.273\\
NiftyReg & 2.751 $\pm$ 1.333\\
ours(next-gen-nn) & 2.224 $\pm$ 0.639 \\
\hline
\end{tabular}
\end{table}

\begin{table}
\centering
\caption{Results of ReMIND2Reg sub-challenge in Learn2Reg 2024.}\label{tab3}
\begin{tabular}{c|c}
\hline
  &  TRE(mm) \\
\hline
Initial & 3.727 $\pm$ 0.714 \\
ConvexAdam-Rigid &  2.773 $\pm$ 1.273\\
NiftyReg & 2.751 $\pm$ 1.333\\
ours(without fusion) & 2.453 $\pm$ 0.613 \\
ours(with fusion) & \textbf{2.224 $\pm$ 0.639} \\
\hline
\end{tabular}
\end{table}

\begin{table}
\centering
\caption{Results of COMULIS3DCLEM sub-challenge in Learn2Reg 2024.}\label{tab2}
\begin{tabular}{|c|c|}
\hline
  &  TRE(LM) \\
\hline
Initial & 50.370 $\pm$ 20.355 \\
ours(next-gen-nn) & 49.609 $\pm$ 20.875 \\
\hline
\end{tabular}
\end{table}

\begin{figure}
\includegraphics[width=\textwidth]{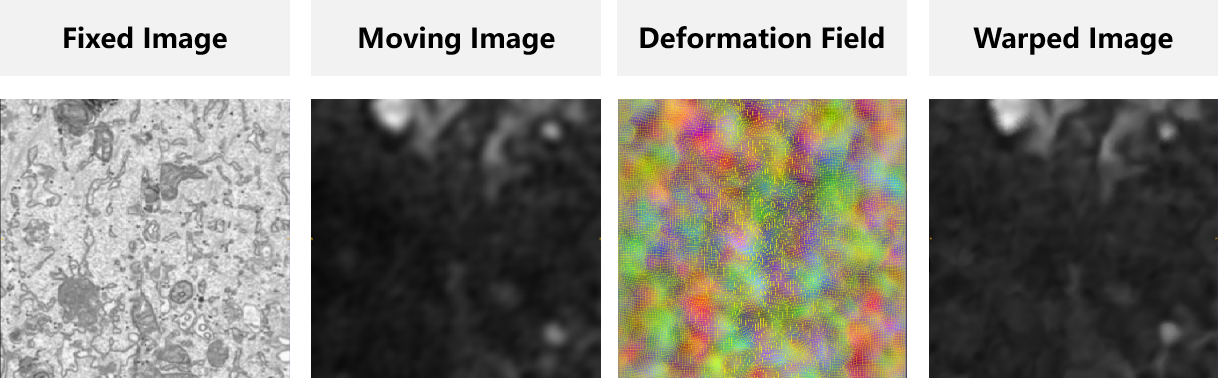}
\caption{Visualization results of COMULIS3DCLEM sub-challenge in Learn2Reg 2024.} \label{fig4}
\end{figure}

\paragraph{COMULIS3DCLEM task.}
Automated registration of multimodal microscope 3D images is a rarely addressed issue in medical image analysis. The COMULIS3DCLEM \cite{ref_article3} sub-challenge aims to align electron microscope (EM) 3D images and light microscope (LM) 3D images of the same cellular region. Since there is no multimodal input of moving images involved in this task, the multimodal fusion part of the method is not used. Since the input image is only $32\times256\times256$, the $n$ of the multilevel pyramidal fusion optimization mechanism in this task is selected as 1, 2, 4, and the weight of each level in Multilevel Weighted Fusion is set as 0.33, 0.33, 0.33.The experimental results are shown in Table~\ref{tab2}. Although the average results are not significant, for some cases, the error of our proposed method can reach 13.882, which is a very competitive registration result for this task. The visualization of the registration for this task is shown in Fig.~\ref{fig4}. In the validation phase, our method ranks in the second place.

\section{Conclusion}
The application of the proposed MCBO method to the Learn2Reg 2024 challenge shows that a multilevel optimization strategy using only a small amount of learning can quickly and accurately achieve the registration between multimodal medical images with large deformations. Especially for the registration of preoperative images and intraoperative images, better results can be obtained by the deep fusion of multimodal moving images. Meanwhile, the method proposed in this paper ranks second in both ReMIND2Reg and COMULIS3DCLEM sub-challenges, which illustrates the generality of the method for multimodal medical image registration.

\begin{credits}

\subsubsection{\ackname} Thanks all the organizers of the MICCAI 2024 Learn2Reg challenge. This work was supported in part by the National Key Research and Development Program of China (grant number 2022YFE0134700), and in part by the National Natural Science Foundation of China (grant number 62221002), and in part by the Fundamental Research Funds for the Central Universities, China.

\end{credits}


\begin{thebibliography}{8}

\bibitem{ref_article4}
Balakrishnan, G., Zhao, A., Sabuncu, M. R., et al.: An Unsupervised Learning Model for Deformable Medical Image Registration. In: 2018 IEEE/CVF Conference on Computer Vision and Pattern Recognition, Salt Lake City, UT, USA pp. 9252-9260, doi: 10.1109/CVPR.2018.00964. (2018)

\bibitem{ref_article5}
Zhang, H., Chen, X., Hu, R., et al.:MemWarp: Discontinuity-Preserving Cardiac Registration with Memorized Anatomical Filters. arXiv preprint arXiv:2407.08093. (2024)

\bibitem{ref_article6}
Zhang, H., Chen, X., Wang, R., et al.:Spatially covariant image registration with text prompts. arXiv preprint arXiv:2311.15607. (2023)

\bibitem{ref_article7}
Chen, X., Diaz-Pinto, A., Ravikumar, N., Frangi, A.F: Deep learning in medical image registration. Progress in Biomedical Engineering \textbf{3}(1), 012003 (2021)

\bibitem{ref_article8}
Chen, X.,  Xia, Y., Ravikumar, N., Frangi, A.F: A deep discontinuity-preserving image registration network. In:  Medical Image Computing and Computer Assisted Intervention--MICCAI 2021: 24th International Conference, Strasbourg, France, September 27--October 1, 2021, Proceedings, Part IV 24, pp. 46-55. (2021)

\bibitem{ref_proc1}
Siebert, H., Hansen, L., Heinrich, M.P: Fast 3D Registration with Accurate Optimisation and Little Learning for Learn2Reg 2021. In:Aubreville, M., Zimmerer, D., Heinrich, M. (eds) Biomedical Image Registration, Domain Generalisation and Out-of-Distribution Analysis. MICCAI 2021. Lecture Notes in Computer Science, vol 13166. Springer, Cham. (2022) https://doi.org/10.1007/978-3-030-97281-3 25

\bibitem{ref_article1}
Heinrich, M., Jenkinson, M.: Mind: Modality Independent Neighbourhood Descriptor for Multi-Modal Deformable Registration. Medical image analysis \textbf{16}(7), 1423-35 (2012)

\bibitem{ref_url1}
ConvexAdam, \url{https://github.com/multimodallearning/convexAdam/tree/main}

\bibitem{ref_proc2}
Heinrich, M.P., Papiez, B.W., Schnabel, J.A., Handels, H.: Non-parametric discrete registration with convex optimisation. In: Ourselin, S., Modat, M. (eds.) WBIR 2014. LNCS, vol. 8545, pp. 51–61. Springer, Cham (2014). https://doi.org/10.1007/ 978-3-319-08554-8 6

\bibitem{ref_article2}
Juvekar, P., Dorent, R., Kögl, F., et al.: The Brain Resection Multimodal Imaging Database (ReMIND). Nature Scientific Data \textbf{11}, 494 (2024). 

\bibitem{ref_article3}
Daniel K., Matouš E., Marie-Charlotte D., et al.: CLEM-Reg: An automated point cloud based registration algorithm for correlative light and volume electron microscopy. bioRxiv 2023.05.11.540445. (2023) doi: https://doi.org/10.1101/2023.05.11.540445

\end{thebibliography}
%

\end{document}